\documentclass{article}

\usepackage{microtype}
\usepackage{graphicx}
\usepackage{subcaption}
\usepackage{booktabs}

\usepackage{hyperref}

\usepackage[accepted]{icml2026}

\usepackage{amsmath}
\usepackage{amssymb}
\usepackage{mathtools}
\usepackage{amsthm}
\usepackage[capitalize,noabbrev]{cleveref}

\usepackage{makecell}
\usepackage{xcolor}

\icmltitlerunning{Accelerating LLM Inference via Vector Index Based Output Embeddings}

\begin{document}

\twocolumn[
  \icmltitle{Accelerating LLM Inference via Vector Index Based Output Embeddings}

  \icmlsetsymbol{equal}{*}

  \begin{icmlauthorlist}
    \icmlauthor{Martin Loretz}{nxai}
    \icmlauthor{Sepp Hochreiter}{nxai,jku}
  \end{icmlauthorlist}

  \icmlaffiliation{nxai}{NXAI, Austria}
  \icmlaffiliation{jku}{
Johannes Kepler University Linz, Austria}
  \icmlcorrespondingauthor{Martin Loretz}{martin.loretz@nx-ai.com}
  \icmlkeywords{LLM Inference Acceleration, Vector index, HNSW, MIPS, Output Embedding}
  \vskip 0.3in
]

\printAffiliationsAndNotice{}

\begin{abstract}
Large output embedding matrices create a significant memory bandwidth bottleneck during autoregressive decoding, especially for compact LLMs with large multilingual vocabularies. We reformulate the output projection followed by top-$k$ token selection as a maximum inner product search over token embeddings and replace the dense vocabulary projection with an HNSW-based vector index. The resulting output head retrieves only a small candidate set of high-scoring tokens and can be integrated into existing decoding pipelines by scattering retrieved logits into a sparse full-vocabulary tensor. On CPU inference with Gemma 3, Llama 3.2, and Qwen 3 models, our method substantially accelerates the output projection and improves end-to-end batch-size-one decoding throughput by up to 82\% for Gemma 3 270M, while preserving generation quality under AlpacaEval evaluation. These results suggest approximate retrieval is a practical alternative to dense output projections in latency-sensitive small-batch decoding.
\end{abstract}

\section{Introduction}

Compact LLMs are increasingly deployed in latency-sensitive edge settings, where decoding is often performed with batch size one. In this regime, the final vocabulary projection can become a dominant bottleneck. This is especially pronounced for recent small models distilled from larger multilingual systems: although their transformer backbones are compact, they often retain vocabularies with 100k–250k tokens. As a result, each decoding step requires streaming a large output embedding matrix from memory merely to select a small set of plausible next tokens.

To address this fundamental inefficiency and accelerate inference, we reformulate the output projection and subsequent top-$k$ sampling step as a single Maximum Inner Product Search (MIPS) problem that can be efficiently solved using approximate retrieval algorithms typically used in vector databases. This approach allows us to directly retrieve the top-$k$ tokens from the final hidden state without the memory bandwidth overhead of loading the entire dense matrix. Unlike structural approximations such as Adaptive Softmax \cite{joulin2017efficient}, which require fundamental architectural changes during pre-training, our approach can be seamlessly integrated into existing models post-training.
It targets the latency-sensitive, small-batch regime typical of interactive local inference, rather than high-throughput server inference where batched GEMM is highly efficient.

In summary, we make the following contributions:
\vspace{-1.0em}
\begin{itemize}
    \setlength{\itemsep}{0pt}
    \setlength{\parskip}{0pt}
    \item We reformulate truncated token sampling as Maximum Inner Product Search (MIPS) over the output embedding matrix.
    \item We introduce a drop-in Hierarchical Navigable Small World (HNSW)-based output head that retrieves candidate logits without evaluating the full vocabulary projection.
    \item We evaluate the method across Gemma 3, Llama 3.2, and Qwen 3 models, showing up to 82\% end-to-end throughput improvement at batch size one with minimal quality degradation.
\end{itemize}

The source code of a reference implementation is available at \url{https://github.com/martinloretzzz/vector-index-embedding}.

\section{Background}

\paragraph{Token Sampling in Large Language Models}
During autoregressive generation, language models \cite{vaswani2017attention} predict a probability distribution over the vocabulary $V$ at each decoding step. While stochastic sampling ensures diverse and natural text generation, drawing from the full distribution often introduces the ``long tail'' problem, where low-probability tokens lead to incoherent outputs. To mitigate this, various truncation methods are utilized, differing primarily in how they determine the distribution cutoff. A prominent approach is top-$k$ sampling \cite{fan2018hierarchical}, which restricts the sampling pool to the $k$ most probable tokens, ensuring only semantically plausible candidates are considered while discarding noisy, low-probability tokens.

\paragraph{Maximum Inner Product Search (MIPS)}
The task of identifying the $k$ tokens that maximize the inner product with the final hidden state $h \in \mathbb{R}^d$ is fundamentally a MIPS problem \cite{shrivastava2014asymmetric}. Given a vocabulary size $|V|$, an exhaustive search requires computing the inner product for every row in the output embedding matrix $W_{out}$. While this brute-force approach guarantees exact results, its $O(|V| \cdot d)$ complexity quickly becomes a computational bottleneck. To achieve sub-linear query times, approximate data structures are utilized to trade marginal losses in retrieval recall for substantial reductions in latency \cite{indyk1998approximate}.

\paragraph{Hierarchical Navigable Small Worlds (HNSW)}
A state-of-the-art algorithm for solving the MIPS problem in high-dimensional spaces is HNSW \cite{malkov2018efficient}. This graph-based approach constructs a proximity graph where nodes represent embedding vectors, and edges connect each node to up to $M$ nearby vectors under the chosen similarity metric. Retrieval is performed by traversing the graph from a predefined entry node toward the query vector $h$ using a greedy search. HNSW utilizes a multi-layered structure where the top layers contain sparse graphs with long-range edges for rapid coarse-grained navigation. As the search descends through the hierarchy, the graphs become increasingly dense, allowing the algorithm to refine the search in the bottom layer. This hierarchical approach enables the search to achieve logarithmic query complexity, $O(\log(|V|) \cdot d)$.

\section{Vector Index Embedding}

In language models, the final hidden state vector $h \in \mathbb{R}^d$ is projected onto the vocabulary space to produce a logit vector $z \in \mathbb{R}^{|V|}$, where $|V|$ is the vocabulary size and $d$ is the hidden dimension. This projection is defined as:
$$z = W_{out} h$$
where $W_{out} \in \mathbb{R}^{|V| \times d}$ represents the output embedding matrix. The logit for an individual token $i$ is the inner product of the hidden state and the token's corresponding embedding vector $w_i$:
$$z_i = \langle w_i, h \rangle$$
Since truncated decoding strategies, such as top-$k$ sampling, only require the largest logits, computing the full vector $z$ introduces an avoidable cost. Treating the rows of $W_{out}$ as a static database of token vectors, selecting the largest logits is exactly a MIPS problem with query $h$.

To solve this search problem efficiently, we evaluated several approximate nearest neighbor algorithms using the FAISS \cite{douze2024faiss} library. Our empirical results demonstrate that HNSW \cite{malkov2018efficient} performs best within this high-dimensional embedding space, achieving an optimal trade-off between retrieval quality and search latency. While alternative algorithms based on clustering \cite{jegou2010product} or the GPU-accelerated CAGRA \cite{ootomo2024cagra} were considered, HNSW was the only candidate that provided sufficient retrieval accuracy to preserve the generative quality of the model.

The index is constructed from the rows of the output embedding matrix using the inner product as the similarity measure. Since these embeddings are static post-training, the index is precomputed and distributed alongside the model checkpoints. This allows us to prioritize index quality over construction time. During inference, the trade-off between throughput and approximation accuracy is controlled through the $ef$ parameter, which dictates the size of the dynamic candidate list maintained during the traversal of the graph. While increasing $ef$ improves recall, smaller values maximize throughput.

Due to the exponentiation in the softmax operation, the final sampling distribution is overwhelmingly dominated by the largest logits. Consequently, the primary risk of an approximate search is failing to retrieve the maximum logit when the model is highly confident, as such an omission causes a large divergence from the ground-truth distribution. We therefore explicitly compute the logits for some special tokens, common punctuation, and structural words to guarantee their inclusion.

While sampling can be performed directly on the retrieved top-$k$ elements, a seamless integration with existing deep learning frameworks requires reconstructing the full logit tensor $z \in \mathbb{R}^{|V|}$ by mapping the retrieved inner products to their vocabulary indices and setting all unretrieved entries to $-\infty$. This approach allows for seamless integration with existing decoding pipelines, enabling the use of standard logit processors, such as top-$p$ \cite{holtzman2019curious} or min-$p$ \cite{nguyen2024turning} sampling, as a second stage.

While our approach drastically reduces memory bandwidth, storing the proximity graph alongside the vector index slightly increases the static memory footprint of the output layer. Assuming 32-bit representations, the memory requirement grows from $|V| \cdot 4d$ bytes for the dense matrix to $|V| \cdot (4d + 8M)$ bytes, representing a 5--10\% overhead.

Overall, this architectural modification reduces the asymptotic complexity of the final projection from $O(d \cdot |V|)$ to approximately $O(d \cdot \log|V|)$. Although the random memory accesses inherent to graph traversal incur a substantially higher constant overhead than contiguous matrix multiplication, these algorithmic efficiency gains ultimately dominate at small batch sizes.

\section{Experiments}

We evaluate our approach on several open-source models from the Gemma 3 \cite{google2025gemma270m}, Llama 3.2 \cite{grattafiori2024llama}, and Qwen 3 \cite{yang2025qwen3} families by replacing their standard output projections with our vector index embedding. Due to the lack of GPU-accelerated HNSW implementations, all evaluations are done on the CPU in single-precision (\texttt{float32}) format. Although weight quantization \cite{frantar2022gptq} and lower-precision data types are standard techniques for reducing memory bandwidth requirements, we focus on algorithmic acceleration, as our approach is orthogonal to these existing methods. We use our own fork of the \texttt{hnswlib} library that optimizes memory access patterns and multithreading for high-dimensional vectors. Benchmarking was performed on a consumer-grade system equipped with a 6-core Intel Core i7-10750H CPU and 32 GB of DDR4 SDRAM, with the CPU frequency locked to 2.60 GHz. All HNSW vector indices were constructed using $M=32$ and $ef_{\text{construction}}=5000$. During text generation, we employed top-$k$ sampling with $k=50$.

We set $ef = 200$ for all primary experiments, as no subjective quality degradation is observed at this threshold. Halving $ef$ to 100 effectively doubles search throughput, offering a viable trade-off for latency-critical applications where minor quality reductions are acceptable (\cref{sec:index-recall}).

\subsection{Output Embedding Benchmarks}

To isolate the performance gains attributable to our approach, we conduct micro-benchmarks comparing the median inference throughput of the HNSW-based output embedding against the standard dense output projection. These evaluations utilize real hidden states extracted from a model processing a Wikipedia article.

\begin{figure}[h]
    \vskip -0.1in
    \begin{center}
        \centerline{\includegraphics[width=\columnwidth]{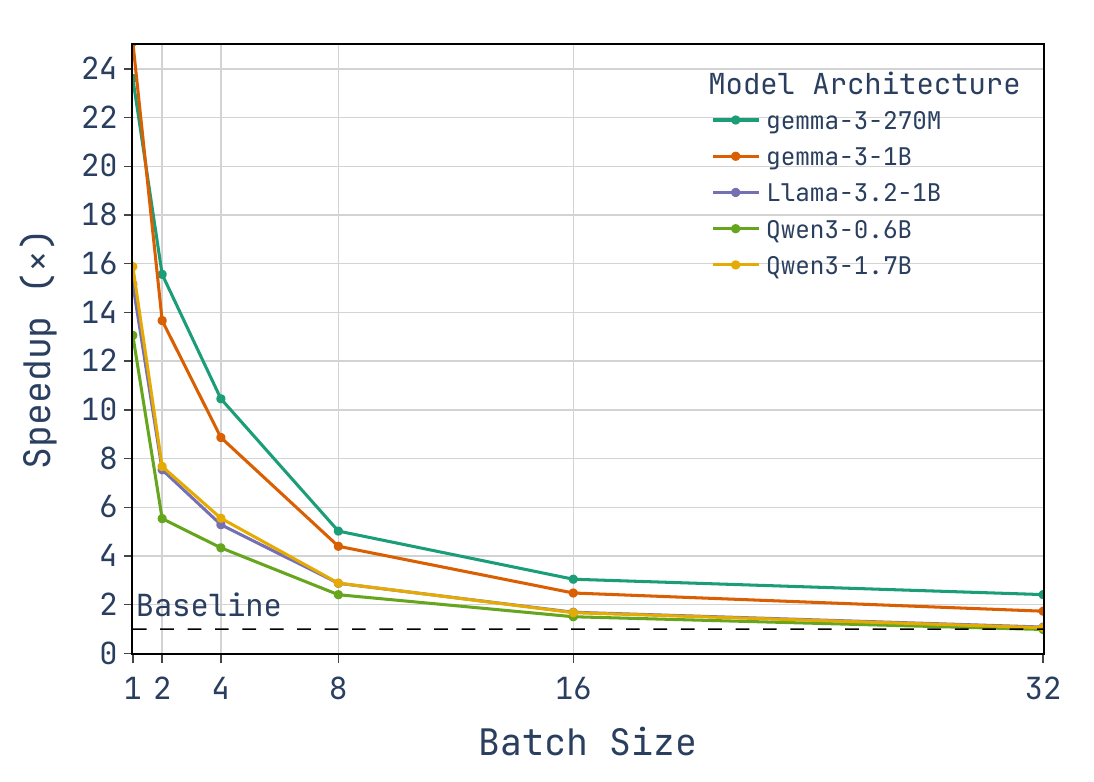}}
        \caption{Relative inference speedup for the output projection at $ef=200$, measured as the ratio of the median throughput of the vector index embedding to the exact matrix multiplication.}
        \label{fig:speedup-head}
    \end{center}
    \vskip -0.1in
\end{figure}

As demonstrated in \cref{fig:speedup-head}, our approach significantly accelerates the output projection relative to exact matrix multiplication, most notably at lower batch sizes. Under these conditions, inference throughput is primarily bottlenecked by memory bandwidth. By loading only a small subset of the embedding matrix, our HNSW formulation effectively bypasses the transfer overhead inherent to evaluating the full projection.

This advantage is most apparent at a batch size of 1, where we observe a speedup exceeding $12\times$ for the Gemma models. As expected, performance gains further scale with vocabulary size, as the 256k-token Gemma models exhibit greater speedups than the smaller-vocabulary Llama (128k) and Qwen (152k) families.

\subsection{Full Model Benchmarks}

Although the vector index is substantially faster than the exact output projection, the aggregate end-to-end performance gain remains fundamentally constrained by the invariant computational overhead of the preceding layers. To quantify this impact, we evaluate the decoding throughput by generating 128-token responses to different short questions across varying batch sizes.

\begin{figure}[h]
    \vskip -0.1in
    \begin{center}
        \centerline{\includegraphics[width=\columnwidth]{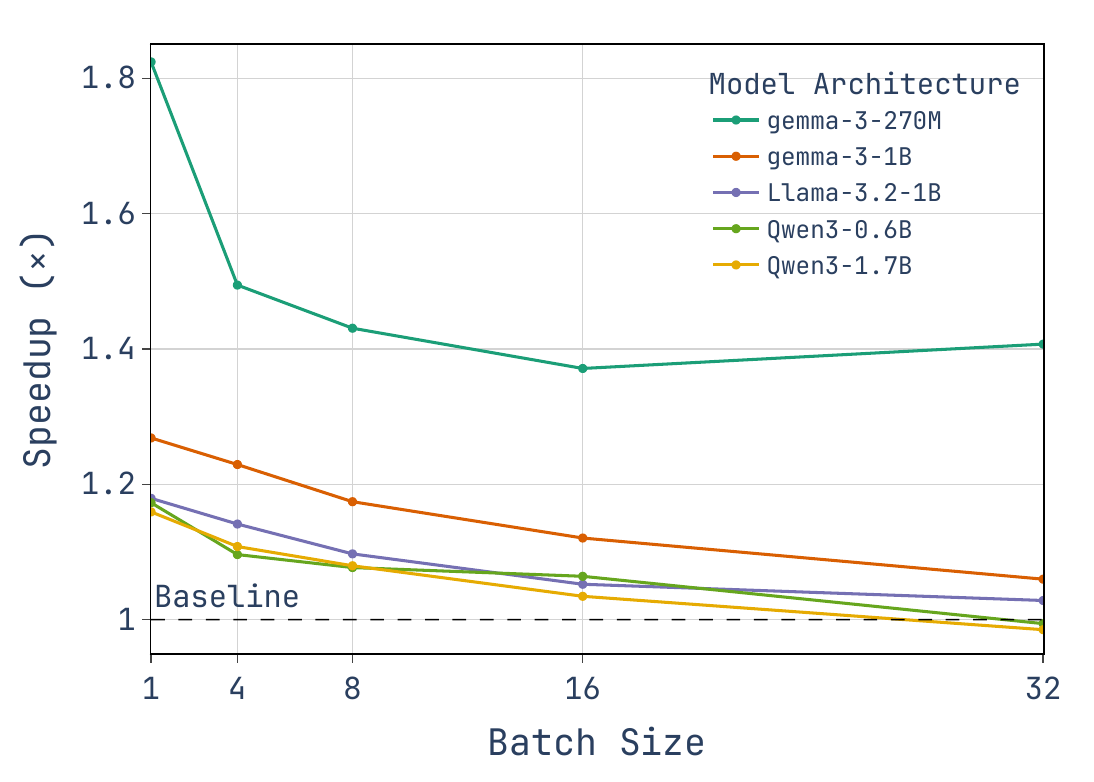}}
        \caption{Relative inference speedup for the full model at $ef=200$, measured as the ratio of the median throughput of the vector index based model to that of the standard baseline.}
        \label{fig:speedup-full}
    \end{center}
    \vskip -0.1in
\end{figure}

As shown in \cref{fig:speedup-full}, the end-to-end inference speedup reaches 82\% during batch-size-one inference for the Gemma~3 270M model, whose embedding matrix alone accounts for approximately 62\% of the total parameter count. Such performance gains underscore the disproportionate computational bottleneck created by large vocabularies in smaller architectures. Notably, this advantage persists at moderate batch sizes, with all evaluated models consistently outperforming the baseline. However, as the batch size exceeds 32, the search overhead offsets these efficiency gains, causing our method to fall behind for some models. Here, the total number of unique tokens retrieved across the batch necessitates loading a substantial fraction of the embedding matrix into memory, thereby negating the memory bandwidth advantages of the vector index. Beyond this crossover point, exact matrix multiplication regains dominance due to the superior cache locality and contiguous memory access patterns of highly optimized GEMM kernels. This reduction in latency is visually illustrated in the profiler traces provided in \cref{fig:profiler-traces} in the Appendix.

\subsection{Index Accuracy}
\label{sec:index-recall}

To evaluate retrieval performance, we measure Recall@2 across varying search depths ($ef$) on 20k tokens from Wikipedia. Here, recall is the percentage of the true top 2 nearest neighbors retrieved by the vector index compared to the exact linear layer. We prioritize Recall@2 as the top two tokens typically capture roughly 70\% of the total softmax probability mass, making them the most critical for preserving generation quality.

\begin{table}[h]
    \center
    \small
    \caption{Retrieval Recall@2 of the HNSW vector index across varying search depths ($ef$).}
    \label{tab:index-recall}
    \begin{sc}
        \begin{tabular}{lccc}
            \toprule
                & \multicolumn{3}{c}{recall@2 (\%)} \\ 
                \cmidrule(lr){2-4}
                Model & $ef=100$ & $ef=200$ & $ef=400$ \\
            \midrule
                Gemma 3 270M    & 97.3 & 99.1 & 99.7 \\
                Gemma 3 1B      & 98.0 & 99.5 & 99.9 \\
                Llama 3.2 1B    & 97.9 & 99.5 & 99.9 \\
                Qwen3 0.6B      & 97.5 & 99.4 & 99.9 \\
                Qwen3 1.7B      & 96.3 & 99.0 & 99.8 \\
            \bottomrule
        \end{tabular}
    \end{sc}
\end{table}

As shown in \cref{tab:index-recall}, recall improves monotonically with $ef$. While the index achieves over 96\% recall at $ef=100$, it introduces a slight degradation in generation quality. Raising the search depth to $ef=200$ pushes recall above 99\%, virtually eliminating divergence from the exact baseline. While $ef=400$ reaches up to 99.9\% recall, the added search overhead justifies our choice of $ef=200$ as the optimal balance between accuracy and speed.

\subsection{Generation Quality}

To evaluate the generative quality of our approach, we employ the LLM-as-a-judge paradigm \cite{zheng2023judging} using the AlpacaEval \cite{alpaca_eval} framework. Specifically, we use GPT-5 Nano \cite{singh2025openai} as the judge to perform a pairwise comparison between the text generated by our vector index model and the standard baseline at $ef=200$. Traditional benchmarks such as MMLU \cite{hendrycks2020measuring} are ineffective in this context, as our method preserves the model's internal representations, resulting in nearly identical scores.

\begin{table}[h]
    \center
    \small
    \caption{Pairwise length-controlled win rates on AlpacaEval, comparing our vector index approach ($ef=200$) to the exact baseline across various model families.}
    \label{tab:alpaca-eval}
    \begin{sc}
        \begin{tabular*}{\columnwidth}{@{\extracolsep{\fill}}lr@{}}
            \toprule
                Model           & \makecell[r]{Length Controlled \\ Win Rate} \\
            \midrule
                Gemma 3 270M	&	48.1	$\pm$ 	0.10	\\
                Gemma 3 1B  	&	45.9	$\pm$ 	0.23	\\
                Llama 3.2 1B	&	49.1	$\pm$ 	0.25	\\
                Qwen3 0.6B  	&	47.5	$\pm$ 	0.26	\\
                Qwen3 1.7B  	&	46.9	$\pm$ 	0.22	\\
            \bottomrule
        \end{tabular*}
    \end{sc}
\end{table}

\newpage

As shown in \cref{tab:alpaca-eval}, the length-controlled win rates for our vector index approach range from 45.9\% to 49.1\% when evaluated against the exact baseline. This self-benchmarking setting is notably demanding because the evaluation is strictly relative, where even minor variations can prompt the judge to favor the baseline. Consequently, win rates approaching 50\% imply the LLM judge could rarely distinguish the optimized model from the exact baseline. While these scores reflect a slight, expected degradation inherent to approximation, the impact on generation quality is negligible. Evaluated alongside the substantial performance gains, this reduction represents a highly favorable trade-off.

\section{Conclusion}

In this work, we introduced a novel framework to accelerate LLM inference by replacing the standard dense output projection with a vector index-based mechanism. This drastically reduces the compute and memory bandwidth requirements for this operation by eliminating the need to evaluate the full projection, while effectively preserving the core generative capabilities of the model. Our results show that for small models with large vocabularies, replacing the dense output layer with a vector index can increase end-to-end inference throughput by as much as 82\% during single-batch inference. These results are particularly relevant for efficient on-device deployment across edge environments such as mobile, automotive, and industrial systems.

Currently, the primary limitation of our approach is its reliance on CPU execution. Although the underlying inner product calculations are parallelizable, the fundamentally sequential nature of graph-based index traversal presents a significant bottleneck for GPU architectures. Consequently, future work will focus on developing GPU-accelerated vector indices and extending this method to support quantized data types.
Finally, the relative benefit may decrease for heavily quantized dense baselines or large batch sizes, where optimized GEMM kernels better amortize memory access.
\newpage

\paragraph{Broader Impact} 
This paper presents work whose goal is to advance the field of Machine
Learning. There are many potential societal consequences of our work, none which we feel must be specifically highlighted here.

\bibliography{paper}

@article{malkov2018efficient,
  title={Efficient and robust approximate nearest neighbor search using hierarchical navigable small world graphs},
  author={Malkov, Yu A and Yashunin, Dmitry A},
  journal={IEEE transactions on pattern analysis and machine intelligence},
  volume={42},
  number={4},
  pages={824--836},
  year={2018},
  publisher={IEEE}
}

@article{jegou2010product,
  title={Product quantization for nearest neighbor search},
  author={Jegou, Herve and Douze, Matthijs and Schmid, Cordelia},
  journal={IEEE transactions on pattern analysis and machine intelligence},
  volume={33},
  number={1},
  pages={117--128},
  year={2010},
  publisher={IEEE}
}

@inproceedings{ootomo2024cagra,
  title={Cagra: Highly parallel graph construction and approximate nearest neighbor search for gpus},
  author={Ootomo, Hiroyuki and Naruse, Akira and Nolet, Corey and Wang, Ray and Feher, Tamas and Wang, Yong},
  booktitle={2024 IEEE 40th International Conference on Data Engineering (ICDE)},
  pages={4236--4247},
  year={2024},
  organization={IEEE}
}

@inproceedings{indyk1998approximate,
  title={Approximate nearest neighbors: towards removing the curse of dimensionality},
  author={Indyk, Piotr and Motwani, Rajeev},
  booktitle={Proceedings of the thirtieth annual ACM symposium on Theory of computing},
  pages={604--613},
  year={1998}
}

@article{vaswani2017attention,
  title={Attention is all you need},
  author={Vaswani, Ashish and Shazeer, Noam and Parmar, Niki and Uszkoreit, Jakob and Jones, Llion and Gomez, Aidan N and Kaiser, {\L}ukasz and Polosukhin, Illia},
  journal={Advances in neural information processing systems},
  volume={30},
  year={2017}
}

@article{shrivastava2014asymmetric,
  title={Asymmetric LSH (ALSH) for sublinear time maximum inner product search (MIPS)},
  author={Shrivastava, Anshumali and Li, Ping},
  journal={Advances in neural information processing systems},
  volume={27},
  year={2014}
}

@article{douze2024faiss,
  title={The Faiss library},
  author={Matthijs Douze and Alexandr Guzhva and Chengqi Deng and Jeff Johnson and Gergely Szilvasy and Pierre-Emmanuel Mazaré and Maria Lomeli and Lucas Hosseini and Hervé Jégou},
  year={2024},
  eprint={2401.08281},
  archivePrefix={arXiv},
  primaryClass={cs.LG}
}

@inproceedings{joulin2017efficient,
  title={Efficient softmax approximation for GPUs},
  author={Joulin, Armand and Ciss{\'e}, Moustapha and Grangier, David and J{\'e}gou, Herv{\'e} and others},
  booktitle={International conference on machine learning},
  pages={1302--1310},
  year={2017},
  organization={PMLR}
}

@article{frantar2022gptq,
  title={Gptq: Accurate post-training quantization for generative pre-trained transformers},
  author={Frantar, Elias and Ashkboos, Saleh and Hoefler, Torsten and Alistarh, Dan},
  journal={arXiv preprint arXiv:2210.17323},
  year={2022}
}

@inproceedings{fan2018hierarchical,
  title={Hierarchical neural story generation},
  author={Fan, Angela and Lewis, Mike and Dauphin, Yann},
  booktitle={Proceedings of the 56th Annual Meeting of the Association for Computational Linguistics (Volume 1: Long Papers)},
  pages={889--898},
  year={2018}
}

@article{holtzman2019curious,
  title={The curious case of neural text degeneration},
  author={Holtzman, Ari and Buys, Jan and Du, Li and Forbes, Maxwell and Choi, Yejin},
  journal={arXiv preprint arXiv:1904.09751},
  year={2019}
}

@article{nguyen2024turning,
  title={Turning up the heat: Min-p sampling for creative and coherent llm outputs},
  author={Nguyen, Minh Nhat and Baker, Andrew and Neo, Clement and Roush, Allen and Kirsch, Andreas and Shwartz-Ziv, Ravid},
  journal={arXiv preprint arXiv:2407.01082},
  year={2024}
}

@article{grattafiori2024llama,
  title={The llama 3 herd of models},
  author={Grattafiori, Aaron and Dubey, Abhimanyu and Jauhri, Abhinav and Pandey, Abhinav and Kadian, Abhishek and Al-Dahle, Ahmad and Letman, Aiesha and Mathur, Akhil and Schelten, Alan and Vaughan, Alex and others},
  journal={arXiv preprint arXiv:2407.21783},
  year={2024}
}

@article{yang2025qwen3,
  title={Qwen3 technical report},
  author={Yang, An and Li, Anfeng and Yang, Baosong and Zhang, Beichen and Hui, Binyuan and Zheng, Bo and Yu, Bowen and Gao, Chang and Huang, Chengen and Lv, Chenxu and others},
  journal={arXiv preprint arXiv:2505.09388},
  year={2025}
}

@misc{google2025gemma270m,
  author = {{Google DeepMind}},
  title = {Introducing Gemma 3 270M: The compact model for hyper-efficient AI},
  howpublished = {Google for Developers Blog},
  year = {2025},
  month = {August},
  url = {https://developers.googleblog.com/en/introducing-gemma-3-270m/}
}

@misc{alpaca_eval,
  author = {Xuechen Li and Tianyi Zhang and Yann Dubois and Rohan Taori and Ishaan Gulrajani and Carlos Guestrin and Percy Liang and Tatsunori B. Hashimoto },
  title = {AlpacaEval: An Automatic Evaluator of Instruction-following Models},
  year = {2023},
  month = {5},
  publisher = {GitHub},
  journal = {GitHub repository},
  howpublished = {\url{https://github.com/tatsu-lab/alpaca_eval}}
}

@article{zheng2023judging,
  title={Judging llm-as-a-judge with mt-bench and chatbot arena},
  author={Zheng, Lianmin and Chiang, Wei-Lin and Sheng, Ying and Zhuang, Siyuan and Wu, Zhanghao and Zhuang, Yonghao and Lin, Zi and Li, Zhuohan and Li, Dacheng and Xing, Eric and others},
  journal={Advances in neural information processing systems},
  volume={36},
  pages={46595--46623},
  year={2023}
}

@article{singh2025openai,
  title={Openai gpt-5 system card},
  author={Singh, Aaditya and Fry, Adam and Perelman, Adam and Tart, Adam and Ganesh, Adi and El-Kishky, Ahmed and McLaughlin, Aidan and Low, Aiden and Ostrow, AJ and Ananthram, Akhila and others},
  journal={arXiv preprint arXiv:2601.03267},
  year={2025}
}

@article{hendrycks2020measuring,
  title={Measuring massive multitask language understanding},
  author={Hendrycks, Dan and Burns, Collin and Basart, Steven and Zou, Andy and Mazeika, Mantas and Song, Dawn and Steinhardt, Jacob},
  journal={arXiv preprint arXiv:2009.03300},
  year={2020}
}
\bibliographystyle{icml2026}

\newpage
\appendix
\onecolumn

\section{Profiler Traces}
\label{sec:profiler-traces}

\begin{figure}[h]
    \centering
    \begin{subfigure}[b]{\textwidth}
        \centering
        \includegraphics[width=\textwidth]{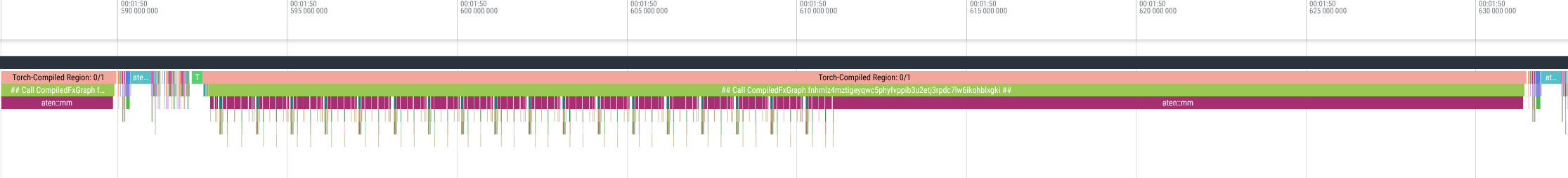}
        \caption{Dense Output Projection}
        \label{fig:trace-baseline}
    \end{subfigure}
    
    \vspace{1.5em}
    \begin{subfigure}[b]{\textwidth}
        \centering
        \includegraphics[width=\textwidth]{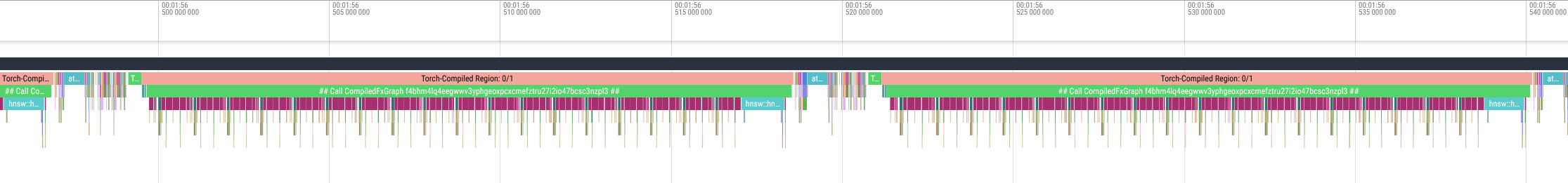}
        \caption{Vector Index Embedding}
        \label{fig:trace-vector}
    \end{subfigure}
    
    \caption{Profiler traces comparing the standard output projection (top) against the vector index embedding (bottom) for the Gemma 3 270M model at a batch size of 1. Both plots share the same timescale and show the final tokens of a 64-token generation. In the baseline trace (a), the per-token latency is heavily dominated by the large matrix multiplications of the output projection. Our proposed vector index embedding (b) eliminates this bottleneck by replacing the dense projection with an HNSW index search. This search takes only a fraction of the time, visible as the narrow blue segment at the end of the decoding step, allowing the model to decode approximately $2\times$ faster than the baseline.}
    \label{fig:profiler-traces}
\end{figure}

\end{document}